\documentclass[11pt,a4paper]{article}

\usepackage{techreport}          
\usepackage{xcolor}
\usepackage{tikz}
\usetikzlibrary{positioning,fit,calc,arrows.meta}

\usepackage[authoryear,sort&compress,round]{natbib}
\usepackage{booktabs}
\usepackage{multirow}
\usepackage{array}
\usepackage{tabularx}
\usepackage{graphicx}
\usepackage{hyperref}
\usepackage{xurl}                 
\usepackage[normalem]{ulem}
\usepackage{cleveref}            
\hypersetup{hidelinks}
\titlespacing*{\section}{0pc}{1.15ex plus2pt minus2pt}{2pt}
\titlespacing*{\subsection}{0pc}{1.0ex plus2pt minus1.5pt}{1pt}
\shorttitle{OmniMed-Jev}                       
\correspondingauthor{cchen@eee.hku.hk, HKU.}
\reportnumber{OmniMed-Jev Technical Report}

\newcommand{\email}[1]{\texttt{#1}}
\newcommand{\orgname}[1]{#1}
\newcommand{\orgaddress}[1]{#1}
\newcommand{\city}[1]{#1}
\newcommand{\country}[1]{#1}

\renewcommand{\trtitlefont}{\normalfont\bfseries\fontsize{14.5}{18}\selectfont}

\title{OmniMed-Jev: Calibrating LVLM Confidence\\for Trustworthy Medical Multimodal Decisions via System One}
\author{Luyao Tang\\ \email{lytang1999@gmail.com}\\[3pt]
  Cheng Chen\\ \email{cchen@eee.hku.hk}\\[6pt]
  {\traffilfont Department of Electrical and Computer Engineering}\\
  {\traffilfont \orgname{The University of Hong Kong}, \orgaddress{\city{Hong Kong SAR}, \country{China}}}}

\newcommand{\medjev}{OmniMed-Jev}
\newcommand{\gen}{Gen-SFT}
\newcommand{\dv}{\ensuremath{\downarrow}}
\newcommand{\up}{\ensuremath{\uparrow}}

\begin{document}
\maketitle

\begin{abstract}
Medical models are judged not only on whether they answer correctly, but on whether the
confidence they report corresponds to how often they are right. Generalist multimodal medical
models have widened what a single model can perceive and solve, yet they still express bounded
decisions such as a diagnosis, a set of findings or a cell count as generated text, so the
probability the model reports is a probability about the next token rather than about the
decision. Motivated by decision-native interfaces such as Jev, we introduce \textbf{\medjev}, which
represents each medical decision as a \textbf{Choice}, \textbf{Noul} or \textbf{Score} decision
over a candidate set supplied at runtime and returns a full distribution over that set:
mutually exclusive classes, binary presence of a finding, or a bounded ordered value. The design
is omni in three respects: the same framework accepts different imaging modalities, covers
different prediction tasks, and expresses them through one candidate-conditioned probability
model, so heterogeneous outputs become comparable probabilities rather than task-specific
strings. Under an interface-controlled comparison against a generative baseline trained on the same backbone, data and schedule, \medjev's reported probabilities track observed correctness far more closely, cutting calibration error by up to an order of magnitude and reliability error by up to two, while point-prediction performance remains comparable and counting is the one family in which the generative baseline stays ahead: making the decision distribution the model's output is not a change of format but what turns the numbers a medical model reports into probabilities that mean what they say. These results
support explicit decision modeling as a way to make reported confidence meaningful within the
tasks evaluated here, and they are not evidence of clinical readiness: the comparison cannot
separate the interface from the training differences that come with it, and we state those
limits with the results. Code is available at \href{https://github.com/lytang63/OmniMed-Jev}{\textcolor{purple}{github.com/lytang63/OmniMed-Jev}}.
\end{abstract}


\section{Introduction}
\label{sec:intro}

Consider two models that make the same diagnosis on the same image and are wrong equally often.
One reports high confidence on the cases it gets wrong; the other lowers its confidence when it is
about to be wrong. Their accuracy is identical, yet only the second is useful for deciding which
cases deserve re-examination, and only the second can be audited against outcomes. Medical AI is
therefore judged twice, on whether its answer is correct and on whether the confidence it reports
corresponds to how often it is right, and this paper is about the second question. In group terms, predictions made at roughly $80\%$ confidence should be correct roughly $80\%$ of the time, and lower confidence should identify a larger share of errors; an individual prediction is either right or wrong, so the requirement is a property of the reported probabilities rather than of a single answer. The mismatch runs in both directions:
\begin{enumerate}\itemsep2pt
\item \textbf{Overconfidence}: the model reports more certainty than its accuracy supports, so errors are accepted with too little scrutiny.
\item \textbf{Underconfidence}: it reports less certainty than its accuracy supports, so cases that could be handled automatically are escalated instead.
\end{enumerate}

Two capabilities are at stake: being accurate, and reporting confidence that can be checked against how often the model is actually right. Generalist medical multimodal large language models have become genuinely generalist in what they
perceive and in what they solve. One model reads histopathology slides, chest radiographs, computed
tomography, fundus photographs, ultrasound and blood-smear microscopy, and answers questions from
naming a tumour subtype to reporting a pleural effusion
\citep{medpalm2,medpalm_expert,llavamed,medflamingo,biomedgpt,medgemma2025,medgemma15}. This
diversity stops at the output. A diagnosis, a set of findings, a cell count and a continuous
measurement are bounded decisions with different structures, and all of them are serialized into
text, so the computation behind the answer is the same conditional token distribution the model
uses to write prose. The probability a token-level model assigns to a string is not directly the
probability of the decision that string encodes, and recovering one from the other is an additional
inference step rather than a property of the model. Alternatives exist, including constrained decoding, explicit candidate scoring and post-hoc calibration, but they all operate on a quantity the model was not trained to produce, and the medical literature has measured the consequence repeatedly: reported confidence overestimates accuracy, calibration varies across specialties, and deferral policies behave unpredictably because the confidence they consume was never the probability of the decision
\citep{bentegeac2025token,savage2025proxies,gastro2026,li2026mindthegap,medvqaoverconf2026,
medabstain2026,jabbour2025selective}.

Typed decision interfaces point at a different arrangement. \emph{Jev motivates the decision
interface; medical multimodality and task heterogeneity motivate its extension.} The interface
makes the candidate set explicit and returns a distribution over it, and three kinds of decision
map naturally onto medical prediction: \textbf{Choice} for mutually exclusive selection such as a
differential diagnosis, \textbf{Noul} for binary presence such as a finding on a radiograph, and
\textbf{Score} for a bounded ordered value such as a cell count or an ejection fraction. Medicine
then adds a requirement of its own. A generalist medical model has to satisfy three requirements at once:
\begin{enumerate}\itemsep2pt
\item \uline{Omni-modality: it accepts what medicine shows it}, different imaging types from histology to fundus photography.
\item \uline{Omni-task: it solves what medicine asks of it}, from naming a class to counting cells.
\item \uline{Omni-decision: it decides in one comparable form}, so that a class, a finding and a grid value are probabilities of the same kind rather than one output format per task.
\end{enumerate}
The third requirement is the methodological claim here, and it is worth separating from the first two: coverage across modalities and tasks is a property of the evaluation suite, whereas a shared candidate-conditioned probability model is what makes heterogeneous decisions expressible, comparable and auditable in the same way. \medjev{} is
therefore our medical implementation and test of the decision-interface idea rather than a medical
copy of a general decision model: it keeps a medical multimodal backbone and adds a shared candidate
scoring mechanism.

This raises an empirical question rather than a claim: if the decision distribution is the model's output object, do the reported probabilities correspond better to observed correctness, and does the model keep its predictive performance? Which families benefit, and which do not? We answer with a controlled comparison: a generative baseline trained on the same backbone, images, adaptation recipe and optimization schedule and evaluated on the same held-out states, so that the comparison isolates how a decision is represented. We hypothesise that a substantial part of the mismatch lies in the output interface; the comparison cannot separate the interface from the denser supervision and the additional decision head that come with it, and we mark that boundary wherever it matters.

Our contributions are:
\begin{itemize}\itemsep2pt\sloppy
\item \textbf{We formulate} confidence--correctness alignment, the correspondence between reported probabilities and observed correctness, as a research question for generalist medical models alongside accuracy.
\item \textbf{We introduce} \medjev, a decision model that is omni in modality, task and decision form: \textbf{Choice}, \textbf{Noul} and \textbf{Score} decisions over candidate sets specified at runtime, with many parallel decisions per state and one shared training objective.
\item \textbf{We evaluate} it under an interface-controlled comparison across four task families and show that the interface buys confidence rather than accuracy: reliability error falls by 9$\times$ to 126$\times$ in three of the four families while point predictions stay comparable, and the one family that favours the generative baseline is a documented property of the ordered-grid representation rather than a limit of the approach.
\end{itemize}

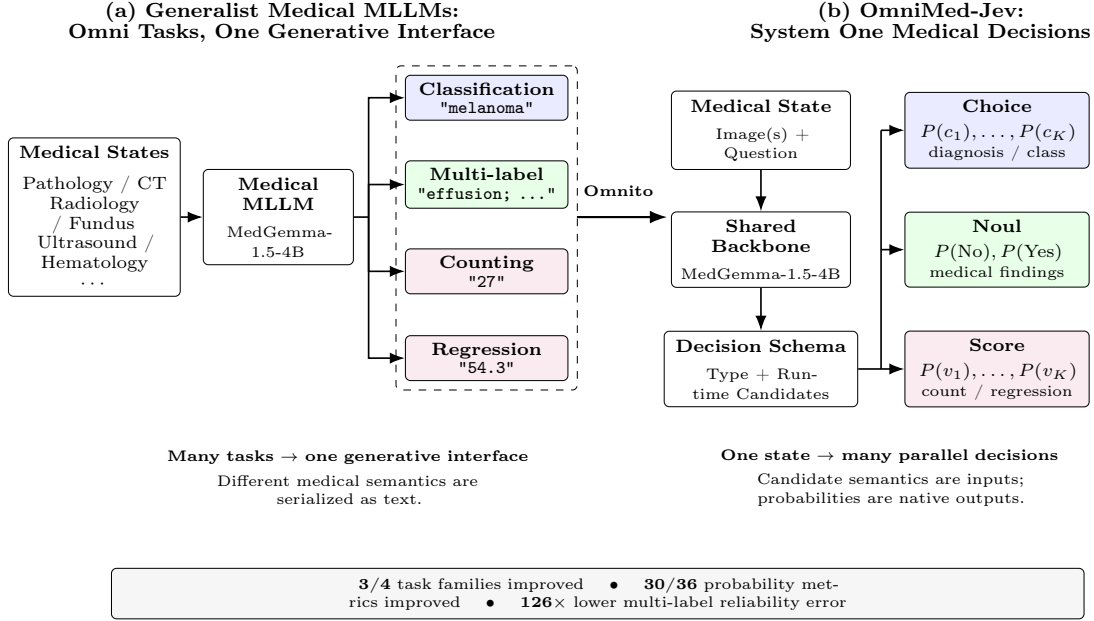
\begin{figure*}[t]
\centering

\resizebox{0.95\textwidth}{!}{%
\begin{tikzpicture}[
    x=0.96cm,
    y=1.50cm,
    >=Latex,
    font=\fontsize{6.7}{7.8}\selectfont,
    box/.style={
        draw,
        rounded corners=2pt,
        align=center,
        inner sep=3.2pt
    },
    smallbox/.style={
        draw,
        rounded corners=2pt,
        align=center,
        inner sep=2.8pt
    },
    choicebox/.style={
        smallbox,
        fill=blue!8
    },
    noulbox/.style={
        smallbox,
        fill=green!9
    },
    scorebox/.style={
        smallbox,
        fill=purple!8
    },
    arrow/.style={
        ->,
        >=Latex,
        line width=0.65pt
    },
    paneltitle/.style={
        font=\fontsize{7.8}{9.0}\selectfont\bfseries,
        align=center
    },
    note/.style={
        font=\fontsize{6.4}{7.4}\selectfont,
        align=center
    },
    tinytext/.style={
        font=\fontsize{5.8}{6.8}\selectfont,
        align=center
    }
]


\node[
    paneltitle,
    text width=7.6cm
] at (3.8,6.15)
{
(a) Generalist Medical MLLMs:\\[-0.2mm]
Omni Tasks, One Generative Interface
};

\node[
    box,
    text width=2.15cm
] (states) at (1.10,4.35)
{
\textbf{Medical States}\\[1.5mm]
Pathology / CT\\
Radiology / Fundus\\
Ultrasound / Hematology\\
$\cdots$
};

\node[
    box,
    text width=1.85cm
] (mllm) at (3.75,4.35)
{
\textbf{Medical MLLM}\\[1mm]
{\fontsize{5.7}{6.5}\selectfont MedGemma-1.5-4B}
};

\node[
    choicebox,
    text width=2.05cm
] (cls) at (6.75,5.45)
{
\textbf{Classification}\\[-0.2mm]
\texttt{"melanoma"}
};

\node[
    noulbox,
    text width=2.05cm
] (multi) at (6.75,4.65)
{
\textbf{Multi-label}\\[-0.2mm]
\texttt{"effusion; ..."}
};

\node[
    scorebox,
    text width=2.05cm
] (count) at (6.75,3.85)
{
\textbf{Counting}\\[-0.2mm]
\texttt{"27"}
};

\node[
    scorebox,
    text width=2.05cm
] (reg) at (6.75,3.05)
{
\textbf{Regression}\\[-0.2mm]
\texttt{"54.3"}
};

\draw[arrow] (states.east) -- (mllm.west);

\coordinate (abranch) at (5.05,4.35);

\draw[
    line width=0.65pt
] (mllm.east) -- (abranch);

\draw[arrow] (abranch) |- (cls.west);
\draw[arrow] (abranch) |- (multi.west);
\draw[arrow] (abranch) |- (count.west);
\draw[arrow] (abranch) |- (reg.west);

\node[
    draw,
    dashed,
    rounded corners=2pt,
    fit=(cls)(multi)(count)(reg),
    inner sep=3.5pt
] {};

\node[
    note,
    text width=5.2cm
] at (4.75,1.95)
{
\textbf{Many tasks $\rightarrow$ one generative interface}\\[0.8mm]
Different medical semantics are\\
serialized as text.
};


\draw[
    ->,
    >=Latex,
    line width=0.85pt
]
(8.05,4.35)
--
node[
    midway,
    above=1.5mm,
    font=\fontsize{5.8}{6.7}\selectfont\bfseries,
    align=center
]
{
Omnito
}
(9.35,4.35);


\node[
    paneltitle,
    text width=7.2cm
] at (13.00,6.15)
{
(b) OmniMed-Jev:\\[-0.2mm]
System One Medical Decisions
};

\node[
    box,
    text width=2.25cm
] (stateb) at (10.70,5.15)
{
\textbf{Medical State}\\[1mm]
{\fontsize{5.8}{6.6}\selectfont Image(s) + Question}
};

\node[
    box,
    text width=2.25cm
] (backbone) at (10.70,4.05)
{
\textbf{Shared Backbone}\\[1mm]
{\fontsize{5.8}{6.6}\selectfont MedGemma-1.5-4B}
};

\node[
    box,
    text width=2.45cm
] (schema) at (10.70,2.95)
{
\textbf{Decision Schema}\\[1mm]
{\fontsize{5.8}{6.6}\selectfont Type + Runtime Candidates}
};

\draw[arrow] (stateb) -- (backbone);
\draw[arrow] (backbone) -- (schema);

\node[
    choicebox,
    text width=2.35cm
] (choice) at (14.10,5.15)
{
\textbf{Choice}\\[0.8mm]
$P(c_1),\ldots,P(c_K)$\\
{\fontsize{5.7}{6.5}\selectfont diagnosis / class}
};

\node[
    noulbox,
    text width=2.35cm
] (noul) at (14.10,4.05)
{
\textbf{Noul}\\[0.8mm]
$P(\mathrm{No}),P(\mathrm{Yes})$\\
{\fontsize{5.7}{6.5}\selectfont medical findings}
};

\node[
    scorebox,
    text width=2.35cm
] (score) at (14.10,2.95)
{
\textbf{Score}\\[0.8mm]
$P(v_1),\ldots,P(v_K)$\\
{\fontsize{5.7}{6.5}\selectfont count / regression}
};

\coordinate (bbranch) at (12.43,2.95);

\draw[
    line width=0.65pt
] (schema.east) -- (bbranch);

\draw[
    line width=0.65pt
] (bbranch) -- ++(0,2.20);

\draw[arrow]
    ($(bbranch)+(0,2.20)$) -- (choice.west);

\draw[arrow]
    ($(bbranch)+(0,1.10)$) -- (noul.west);

\draw[arrow]
    (bbranch) -- (score.west);

\node[
    note,
    text width=5.2cm
] at (12.55,1.95)
{
\textbf{One state $\rightarrow$ many parallel decisions}\\[0.8mm]
Candidate semantics are inputs;\\
probabilities are native outputs.
};


\node[
    draw,
    rounded corners=2pt,
    fill=gray!7,
    text width=13.2cm,
    align=center,
    inner sep=3.5pt,
    font=\fontsize{6.2}{7.2}\selectfont
] at (8.35,0.88)
{
\textbf{3/4} task families improved
\quad $\bullet$ \quad
\textbf{30/36} probability metrics improved
\quad $\bullet$ \quad
\textbf{126$\times$} lower multi-label reliability error
};

\end{tikzpicture}%
}

\caption{
\textbf{OmniMed-Jev rethinks medical MLLMs as System One decision models.}
(a) Generalist medical MLLMs already span heterogeneous modalities and tasks,
yet bounded predictions with fundamentally different semantics are serialized
through the same autoregressive text interface.
Classification, multi-label recognition, counting, and regression therefore share
the same generative output mechanism despite requiring distinct decision semantics.
(b) OmniMed-Jev replaces unnecessary generation with typed probabilistic decisions
over runtime-specified candidate spaces.
The shared medical backbone directly produces Choice, Noul, and Score distributions,
allowing one medical state to support multiple parallel decisions.
When a medical output is inherently a decision, OmniMed-Jev decides directly rather
than first generating that decision as text.
}
\label{fig:overview}

\end{figure*}

\section{Related Work}
\label{sec:related}

\subsection{Generalist medical multimodal models}
\label{sec:related-med}
The generalist turn in medical AI has widened perception and task coverage without changing the
output interface. Clinical language models first showed that medical knowledge can be elicited
to expert-level accuracy on multiple-choice questions \citep{medpalm2,medpalm_expert}, and
multimodal instruction tuning carried the recipe to images: medical visual assistants
\citep{llavamed,medflamingo}, unified biomedical pretraining \citep{biomedgpt},
chest-radiograph and radiology-report systems \citep{chexagent,maira2,llavarad}, and open
medical backbones evaluated across modalities with shared weights
\citep{medgemma2025,medgemma15}. Benchmarks grew in parallel, from curated VQA collections to broad examinations and clinician-written rubrics \citep{slake,vqarad,medmnistv2,chestxdet,medmcqa,medqa,healthbench}. What this literature does not unify is how an answer is represented: classification, findings, counts and measurements are all serialized into strings and compared against a reference, so neither the objective nor the evaluation asks the model to distribute probability mass over a finite, auditable set of options.

\subsection{Uncertainty and calibration in medicine}
\label{sec:related-cal}
Measurements of medical model confidence converge on one result: what a model reports about its own certainty carries little information about whether it is right, whether verbalized or read from token probabilities
\citep{bentegeac2025token,savage2025proxies,gastro2026,li2026mindthegap}, and multimodal
medical models inherit the deficit \citep{medvqaoverconf2026,medvqacalib2025,mvcbenc2026}. The cause is structural: instruction tuning and preference optimization raise accuracy while inflating confidence \citep{xiong2024can,huang2024relying}, so the step that makes a model useful damages its probability. Both standard responses keep the interface intact: post-hoc calibration rescales the model's output
\citep{guo2017calibration,zhao2021calibrate,xie2024ats}; calibration-aware objectives train it
to emit a calibrated confidence
\citep{damani2026rlcr,baniharouni2026rewarding,confTuner2025}, which is then verbalized as
text. The deferral, human-factors and regulatory literature inherits that object
(Appendix~\ref{app:related}).

\subsection{From language likelihood to decision probability}
\label{sec:related-decision}
The machinery that consumes a probability is old and well developed: the reject option, selective classification and conformal prediction take the probability as given and optimize what to do with it \citep{chow1970,geifman2017selective,angelopoulos2023conformal}. In medicine the required calibrated probability is often unavailable: abstention is close to orthogonal to accuracy \citep{medabstain2026,medqabstain2026}, deferral relocates error
rather than removing it \citep{jabbour2025selective}, and cascading on the same signal buys
little \citep{fanconi2025cascade}. A recent wave of decision-native models removes the generation step entirely, making the candidate set part of the input and returning a distribution over it without emitting text \citep{jev2026typesafe,jevjudge2026,decisionmodels2026,jevscale2026,jevmem2026,openjev2026,reflex2026,thisthatmodel2026,visualjev2026}; the framing is old \citep{kahneman2011thinking}. The one controlled study of how such a model binds answers to options shows
that swapping which rubric is attached to which option name changes a substantial fraction of
its answers while the type-error rate stays at zero \citep{typesafeerrorfree2026}, and the single
medical evaluation to date judges report factuality without measuring calibration
\citep{jevradiology2026}.

The field therefore diagnoses the problem and proposes an interface, but nothing connects the two under control: confidence studies repair generative confidence post hoc, calibration-aware objectives still verbalize the probability, and decision-native models report cost and accuracy rather than calibration. This paper supplies that controlled comparison.

\section{OmniMed-Jev}
\label{sec:method}

\subsection{From generation to decisions}
\label{sec:method-gen}
A generative medical MLLM factorizes its answer as a product of token probabilities,
$p_{\mathrm{LM}}(y \mid S, Q) = \prod_{t} p(y_t \mid y_{<t}, S, Q)$, where $S$ is a multimodal
medical state and $Q$ a question. Under that interface a diagnosis, a set of findings, a count
and a continuous measurement are all sequences of tokens, and any notion of a decision
probability has to be recovered from the factorization after the fact. \medjev{} replaces the
answer string with a decision, $D_j = (Q_j, T_j, C_j)$ over a candidate space
$C_j = \{c_{j1}, \dots, c_{jK_j}\}$, where $T_j$ is the decision type, and trains the model to
return a distribution over that space,
\begin{equation}
\label{eq:decision-dist}
p_\theta\!\left(c \mid S, D_j\right), \qquad c \in C_j, \qquad
\sum_{c \in C_j} p_\theta(c \mid S, D_j) = 1 .
\end{equation}
The output is thus a distribution over an enumerable set rather than a string; a point
prediction, when one is needed, is a deterministic function of that distribution, which we keep
as the primary object because it is what a downstream decision rule consumes. Medical decisions differ in structure, and the interface reflects this with three types, each
defined by what a decision rule needs from it. \textbf{Choice} is a mutually exclusive selection: the candidates are the classes, as in a differential diagnosis, and what the rule needs is the probability of each class rather than only the name of the winner.
\textbf{Noul} is binary presence: one decision per finding per state, as in whether a pleural
effusion is present on a radiograph, and each finding needs its own probability because a rule that
escalates findings acts on them one by one. \textbf{Score} is a bounded ordered value: the
candidates are the admissible values in order, as in a cell count or an ejection fraction, and a
distribution over that grid is what makes an expectation and an interval available; because position
defines the order, Score candidate sets are never permuted. Free-form report generation and
bounding-box detection are outside the scope of this paper: span decisions for boxes were removed on
both models because the coordinate representation did not localize findings in our implementation
(Appendix~\ref{app:suite}), and unbounded text has no candidate set to score. Neither exclusion is a
claim about what the interface could express.

\subsection{Candidate-conditioned shared decision scoring}
\label{sec:method-score}
Omni-modeling is the methodological contribution, and the commitment behind it is that candidates are \emph{input semantics} rather than indices of a fixed classifier. All three decision types are scored by one mechanism, so a class, a finding and a grid value are represented alike and their probabilities are comparable. A decision block is rendered into the context: the question
followed by the candidate options, each terminated by a marker token, with a second marker
closing the decision. The model reads this sequence with the image interleaved, and a shared
scoring head consumes the hidden state at the candidate marker together with the state
representation,
\begin{equation}
\label{eq:scorer}
z_{jk} = f_\phi\!\left(h_{\mathrm{state}},\, h_{c_{jk}}\right), \qquad
p_{jk} = \frac{\exp(z_{jk})}{\sum_{l} \exp(z_{jl})} .
\end{equation}
Because the head reads hidden states at marked positions, no projection to the vocabulary is required, and because the candidate text is in the context, the model must read the option it is scoring instead of learning a positional label map. Training reshuffles Choice and Noul candidates every epoch, while Score candidates stay in place because their position encodes their value; the generative baseline receives the equivalent augmentation under the same schedule, since comparing an interface that ignores option order against one that exploits it would confound the comparison.

\subsection{One state, many decisions}
\label{sec:method-many}
A medical state usually carries several decisions at once: a chest radiograph is annotated with a set of findings, a blood smear with a total count and cell-type counts. The interface encodes the state once and returns all of its decisions in one forward pass,
\begin{equation}
\label{eq:many}
S \;\longrightarrow\; \{D_1, \dots, D_M\}, \qquad
\left\{p_\theta(\cdot \mid S, D_1), \dots, p_\theta(\cdot \mid S, D_M)\right\}.
\end{equation}
Under the generative interface the same information is serialized into one string or elicited with separate prompts. For training, the annotations are factorized according to their decision semantics rather than collapsed into one answer per image, so the supervision is denser even though no new labels are collected.

\subsection{Unified decision learning}
\label{sec:method-loss}
All three decision types are trained with one objective, the cross-entropy between the model
distribution and the target distribution induced by the annotation for each decision,
\begin{equation}
\label{eq:loss}
\mathcal{L}(\theta) = -\frac{1}{M} \sum_{j=1}^{M} \sum_{k=1}^{K_j} q_{jk} \log p_{jk},
\end{equation}
where $q_{jk}$ is the target distribution over $C_j$ for decision $j$ and $M$ the number of
decisions in the batch. Adaptation proceeds in two stages on top of the frozen backbone: the
decision head is trained first, and then the head is trained together with low-rank adapters.
The generative baseline uses the same backbone, adapter configuration and number of optimization
steps with the corresponding next-token objective. Hyperparameters, parameter counts and
compute are reported in Section~\ref{sec:exp-sft} and Appendix~\ref{app:train}.

\begin{table}[t]
  \centering
  \caption{Point-prediction performance of the two models on the held-out suite ($n{=}200$ for
  classification, multi-label and regression; $n{=}126$ for counting). Task metrics are computed
  by the task evaluator on folded predictions; the two models share the backbone, training images
  and main training budget, and their output objectives differ (Section~\ref{sec:exp-sft}).
  Multi-label F1 is a macro average; under micro averaging the models exchange places. Paired
  bootstrap intervals and McNemar and sign tests are in Appendix~\ref{app:train}.}
  \label{tab:pred}
  \footnotesize
  \setlength{\tabcolsep}{5pt}
  \begin{tabular}{@{}llrrr@{}}
    \toprule
    Task family & Metric & \gen{} & \medjev{} & $\Delta$ \\
    \midrule
    Classification & Balanced accuracy $\up$ & 0.6469 & \textbf{0.6892} & $+0.0422$ \\
    \midrule
    Multi-label    & F1 $\up$                & 0.5449 & \textbf{0.5618} & $+0.0170$ \\
                   & Precision $\up$         & 0.5679 & \textbf{0.5950} & $+0.0271$ \\
                   & Recall $\up$            & 0.5498 & \textbf{0.5602} & $+0.0104$ \\
    \midrule
    Counting       & MAE $\dv$               & \textbf{2.1429} & 2.5159 & $+0.3730$ \\
    \midrule
    Regression     & MAE $\dv$               & 3.3400 & \textbf{2.9488} & $-0.3912$ \\
    \bottomrule
  \end{tabular}
\end{table}

\section{Experiments}
\label{sec:exp}

\subsection{An omni-task, omni-modality suite}
\label{sec:exp-suite}
Table~\ref{tab:suite} in Appendix~\ref{app:suite} summarizes the evaluation suite. Four task families instantiate the three decision types and are drawn from fifteen source datasets covering histology, pathology, computed tomography, haematology, dermatology, ophthalmology, chest radiography, ultrasound and fundus photography. Classification pools ten MedMNIST collections into a single choice problem
over their organ and tissue classes; multi-label recognition asks one presence decision per
chest-radiograph finding; counting predicts cell counts in blood-smear microscopy; and
regression predicts left-ventricular ejection fraction from cardiac ultrasound and a retinal
score from fundus photographs. The suite is deliberately heterogeneous: it contains families with overlapping label semantics across modalities, families with many parallel decisions per state, and families whose candidate space is a large ordered grid. The tasks were selected and the splits constructed by us rather than adopted from a published benchmark, so that neither model can profit from a generalist model having been tuned against a widely used evaluation; what the comparison reflects is the two output interfaces, not two levels of familiarity with a known test set. Splits and candidate grids are in Appendix~\ref{app:suite}.

\subsection{Generative baseline and matched settings}
\label{sec:exp-sft}
The comparison is interface-controlled: the two models share the instruction-tuned medical backbone, the training images, the preprocessing, the low-rank adapter configuration and the main training budget, and differ in their output objective and in the training implementation it requires. The generative baseline answers by producing the answer text under the
corresponding next-token objective. \medjev{} adapts in two stages: the decision head is trained
first, and then the head is trained together with the adapters. The head
carries $1{,}057{,}410$ parameters; the adapted stage trains $39{,}555{,}202$ of the backbone's
$2{,}529{,}778{,}162$ parameters, $1.564\%$. Both models are evaluated at step 2424, \medjev's best-validation checkpoint; full hyperparameters and hardware are in Appendix~\ref{app:train}.

Two further differences follow from the interface rather than from tuning: because annotations are factorized into one target per decision, \medjev{} sees more supervision per image than a baseline that emits a single answer, and its decision head is an additional trained component. Appendix~\ref{app:limits} lists them with the other respects in which the two models are not like-for-like.

\subsection{Metrics}
\label{sec:exp-metrics}
Four families of metrics are reported, each measuring a property a medical decision rule consumes: calibration, the separation of reliability from resolution, selective behaviour, and decision-space coverage. Their definitions are in Appendix~\ref{app:metrics} and, for the quantities of Table~\ref{tab:quality}, in Section~\ref{sec:res-cal}.

\subsection{Recovering probabilities from the generative baseline}
\label{sec:exp-recover}
A generative model does not emit a distribution over candidates, so the comparison requires a
reconstruction. We render the same decision block for the baseline, score each candidate by the
summed log-probability of its continuation under the language model, and normalize over the
candidate set,
\begin{equation}
\label{eq:recover}
\tilde{p}(c_{jk}) = \frac{\exp \sum_{t} \log p_{\mathrm{LM}}\!\left(y^{(jk)}_{t} \mid y^{(jk)}_{<t}, S, D_j\right)}
{\sum_{l} \exp \sum_{t} \log p_{\mathrm{LM}}\!\left(y^{(jl)}_{t} \mid y^{(jl)}_{<t}, S, D_j\right)} ,
\end{equation}
where $y^{(jk)}_{t}$ denotes the tokens of candidate $c_{jk}$ in the same decision block, so a candidate is scored as a whole continuation rather than by its opening token. The reconstruction is generous, since the baseline receives the same candidate set and prompt as \medjev, and it remains a reconstruction: the continuation need not be the text the model would have produced. Appendix~\ref{app:recover} gives the estimator, the alternatives we tried and its failure modes.

\section{Results}
\label{sec:res}

\subsection{Accuracy is not where the interface changes the model}
\label{sec:res-pred}
Table~\ref{tab:pred} reports the task metrics. The two models land within a few points of each other on every family, and the differences that appear to favour decision modeling do not exclude a null difference: the classification, multi-label and regression gaps are not distinguishable from zero at $95\%$, the multi-label ordering reverses under pooled micro averaging, and the tally is checkpoint-dependent. The one difference distinguishable from zero is the counting MAE, and it favours the generative baseline. The point-prediction comparison therefore establishes no advantage for either representation, which is exactly the result the rest of the paper needs: the question is not whether the decision interface predicts better, but what the probabilities it reports are worth.

\begin{figure}[t]
  \centering
  \includegraphics[width=0.90\linewidth]{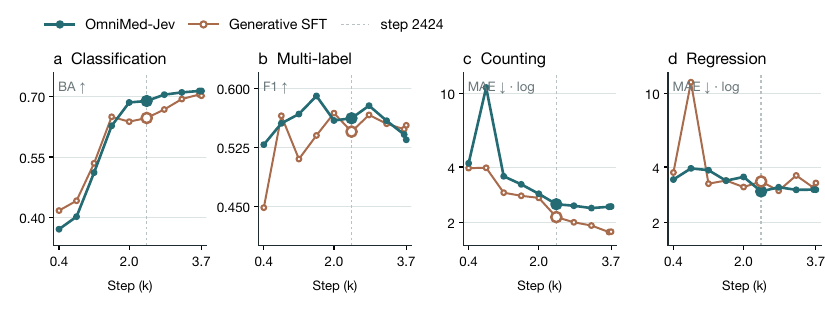}
  \caption{\textbf{Learning dynamics across four task families.} All ten matched test checkpoints
  are shown; the dashed line and enlarged markers mark the comparison step. Counting and
  regression use logarithmic MAE axes to retain early outliers without hiding later differences.
  The interfaces differ in trajectory as well as in endpoint: the generative baseline leads early
  in classification and is overtaken after step 1616, while in regression it passes through a
  large mid-training excursion that \medjev{} does not exhibit.}
  \label{fig:dynamics}
\end{figure}

\subsection{Calibration and discrimination}
\label{sec:res-cal}
The reported probability in this section is the probability that the model's top candidate is correct: of the class it selects for a Choice decision, of the finding it considers present for a Noul decision, and of the annotated value for a Score decision. It is a probability about the decision, not the probability that a disease is present or that a measurement is clinically acceptable; those are different quantities, defined by a downstream rule, and the metrics below measure the one the model reports. Making that distribution a native output makes calibration directly trainable and measurable rather than a post-hoc repair, and the answer to whether it is in fact calibrated is the central empirical claim of this paper.

\textbf{Calibration.} Expected calibration error bins decisions by reported confidence and
averages the gap between confidence and accuracy,
\begin{equation}
\label{eq:ece-main}
\mathrm{ECE} = \sum_{b=1}^{B} \frac{n_b}{n}\,
\bigl|\mathrm{acc}(b) - \mathrm{conf}(b)\bigr| ,
\end{equation}
where a decision's confidence is the probability of its top candidate. The quantity answers the
question a clinician asks of a reported probability: when the model says it is right with
probability $p$, is it right about $p$ of the time? The generative baseline is badly miscalibrated everywhere: its stated confidence is off by $0.26$ on classification, $0.30$ on multi-label recognition, $0.42$ on counting and $0.13$ on regression, so a threshold placed on that number is not a probability that the decision is correct. \medjev{} cuts the error to $0.08$, $0.01$, $0.04$ and $0.09$ in the same four families, a reduction of $3\times$ to $26\times$, and its reported probabilities sit close to the rate at which its decisions are actually correct. \textbf{This is the paper's central empirical claim}: the same backbone, trained to output the decision distribution instead of the answer string, reports numbers that mean what they say.

\begin{figure}[t]
  \centering
  \includegraphics[width=0.8\linewidth]{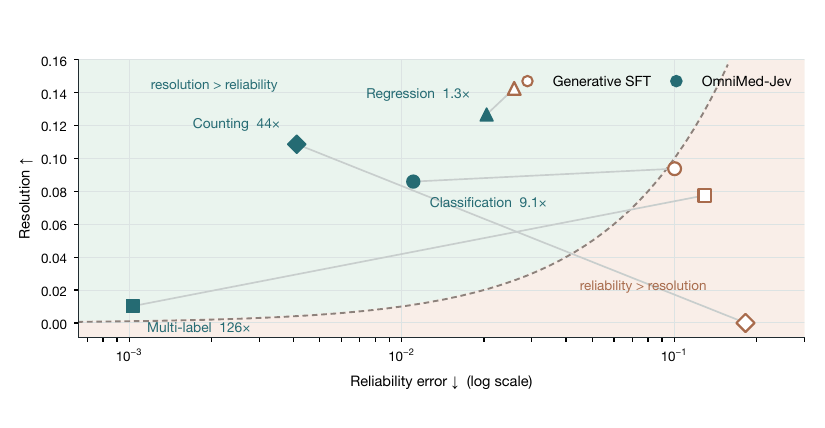}
  \caption{\textbf{Break-even boundary for top-label probability skill.} The dashed boundary is
  $\text{resolution} = \text{reliability}$ in the Murphy decomposition; green marks
  $\text{resolution} > \text{reliability}$ and pale rust the opposite. All four \medjev{}
  families are on the green side; the generative baseline crosses to the rust side in
  classification, multi-label recognition and counting. The boundary is relative to each model's
  own top-label correctness rate, not a clinical decision threshold.}
  \label{fig:skill}
\end{figure}

\textbf{Reliability versus resolution.} The Brier score decomposes into a calibration term, a
discrimination term and an irreducible constant,
\begin{equation}
\label{eq:murphy-main}
\mathrm{BS} = \underbrace{\sum_b \tfrac{n_b}{n}\bigl(\mathrm{conf}_b-\mathrm{acc}_b\bigr)^2}_{\text{reliability}}
- \underbrace{\sum_b \tfrac{n_b}{n}\bigl(\mathrm{acc}_b-\bar{y}\bigr)^2}_{\text{resolution}}
+ \underbrace{\bar{y}(1-\bar{y})}_{\text{uncertainty}} ,
\end{equation}
with $\bar{y}$ the base correctness rate. Reliability error asks whether the reported numbers are
probabilities; resolution asks whether the ranking separates correct from incorrect decisions. The two move differently, and both directions have to be reported. The generative baseline passes the second test more often than the first: its reliability error exceeds its resolution in classification, multi-label recognition and counting (Figure~\ref{fig:skill}), so it orders cases usefully while its reported numbers overstate how often it is right. \medjev{} inverts the pattern decisively, cutting reliability error by $9\times$ on classification, $44\times$ on counting and $126\times$ on multi-label recognition while resolution stays comparable in classification and regression: the same ordering is expressed in a quantity a downstream rule can consume.

\begin{table}[t]
  \centering
  \caption{Decision quality of the two models. Each cell gives \gen{} / \medjev; lower is better
  for ECE, reliability and AURC, higher for resolution. Reliability and resolution are the Murphy
  decomposition of the top-label decision. The generative baseline's distribution is the
  reconstruction of Eq.~\eqref{eq:recover} rather than a trained output, and \medjev's values are
  from the same step. Counting is not a like-for-like sample size;
  Section~\ref{sec:res-negative} reports the matched comparison.}
  \label{tab:quality}
  \footnotesize
  \setlength{\tabcolsep}{4pt}
  \begin{tabular}{@{}lcccc@{}}
    \toprule
    Task family ($n$) & ECE $\dv$ & Reliability $\dv$ & Resolution $\up$ & AURC $\dv$ \\
    \midrule
    Classification (200) & 0.2589 / \textbf{0.0806} & 0.1002 / \textbf{0.0110} & \textbf{0.0938} / 0.0860 & 0.2020 / \textbf{0.1351} \\
    Multi-label (2{,}795) & 0.2981 / \textbf{0.0114} & 0.1294 / \textbf{0.0010} & \textbf{0.0776} / 0.0102 & 0.2474 / \textbf{0.0126} \\
    Counting (201 / 126) & 0.4204 / \textbf{0.0408} & 0.1824 / \textbf{0.0041} & \textbf{0.0000} / \textbf{0.1087} & 1.0000 / \textbf{0.3890} \\
    Regression (200) & 0.1267 / \textbf{0.0913} & 0.0258 / \textbf{0.0205} & \textbf{0.1426} / 0.1268 & \textbf{0.2272} / 0.2280 \\
    \bottomrule
  \end{tabular}
\end{table}

The multi-label family is where the interface pays off most, because there the decisions are many and independent: reliability error falls from $0.1294$ to $0.0010$, a factor of $126$, and the area under the selective risk curve from $0.2474$ to $0.0126$, a factor of $20$. A radiograph carries fourteen presence decisions, and \medjev's per-finding probabilities are close to the rate at which each finding is correct, so a rule that escalates a finding above a fixed probability now behaves the way its specification says. Its ranking of findings is flatter than the baseline's ($0.0776$ against $0.0102$), which costs nothing operationally here: the baseline's ordering is expressed in numbers that overstate correctness by a quarter, and its selective risk is twenty times higher as a result. Regression is the family where the interface adds least, improving calibration ($0.1267$ to $0.0913$) without changing selective behaviour.

\subsection{Does confidence identify errors?}
\label{sec:res-selective}
Selective prediction tests whether the reported confidence can find the errors: ordering decisions by confidence and answering only the most confident fraction gives a risk-coverage curve, whose error at coverage $c$ and area are
\begin{equation}
\label{eq:aurc-main}
\mathrm{risk}(c) = \text{error among the } c \text{ most confident decisions}, \qquad
\mathrm{AURC} = \int_0^1 \mathrm{risk}(c)\, dc ,
\end{equation}
and the quantity a deferral policy consumes is whether low confidence marks the decisions worth handing to a clinician at the stated error level.

\begin{figure}[t]
  \centering
  \includegraphics[width=0.9\linewidth]{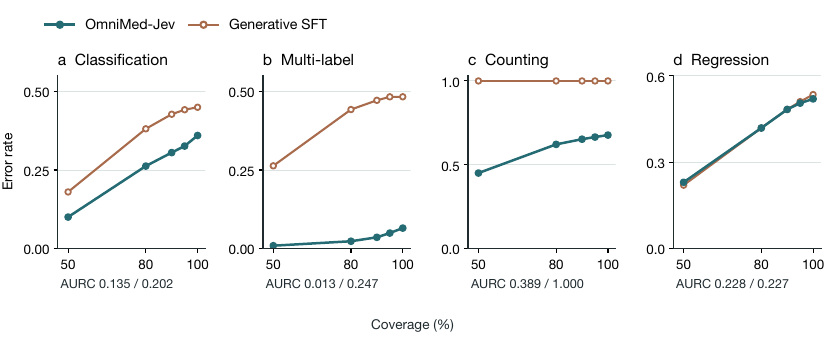}
  \caption{\textbf{Selective prediction at reported coverage levels.} Points show top-label error
  when the model answers the most confident 50\%, 80\%, 90\%, 95\% and 100\% of decisions;
  lines connect only those reported aggregates, not a reconstructed full curve. Panel AURC pairs
  are \medjev{} / \gen. For counting the baseline's reconstruction orders no decision correctly,
  so its curve is flat at an error of $1.0$, which reflects the reconstruction of
  Section~\ref{sec:exp-recover} rather than generative confidence in general.}
  \label{fig:risk}
\end{figure}

Ranking by \medjev's probabilities and retaining the most confident half leaves a much lower error
rate than ranking by the baseline's reconstructed confidence at the same coverage, and the gap
widens as coverage increases (Figure~\ref{fig:risk}); on multi-label recognition the area under the curve falls from $0.2474$ to $0.0126$, a twenty-fold reduction in the error a deferral policy carries at every coverage level.

Counting is the case in which the curve measures the reconstruction rather than the interface. The baseline's reconstructed distribution places no count on the annotated value in any of the $126$ counting decisions, so its ordering is uninformative and its curve is flat at an error of $1.0$, which the AURC value follows from. We read this as a failure of the reconstruction when the answer is a value on a large ordered grid, not as evidence that a generative model cannot order counting decisions at all.

The same distributions support shortlists. $\mathrm{DDx}@0.9$ is the number of top-ranked candidates needed to carry $0.9$ of the probability mass, and it shrinks in every family (Appendix~\ref{app:full}), which says that the distribution is more concentrated; whether the annotated candidate lies inside a shortlist of a given size is a separate quantity, and the two should not be conflated. For a differential diagnosis the difference is between a bounded shortlist whose length is part of the model's output and a ranked list whose length the clinician has to choose.

\subsection{Manipulating the decision space at inference time}
\label{sec:res-probes}
Because candidates are input semantics, the decision space can be manipulated at inference time without retraining, and the effect depends on the family (Table~\ref{tab:probes}): permuting the candidate order leaves classification answers almost unchanged (top-1 agreement $0.90$) while counting and regression answers change substantially ($0.42$ and $0.67$), consistent with a model sensitive to the semantics of the options it reads rather than to their positions.
Restricting the candidate set to a short sub-list leaves accuracy far above chance ($0.87$ for a
two-option list against a chance level of $0.50$), and removing the annotated candidate makes the
model select a plausible runner-up ($0.88$) rather than collapse. None of these operations has a
counterpart under the generative interface without adding a separate scoring mechanism
(Appendix~\ref{app:full}).

\subsection{The counting boundary}
\label{sec:res-negative}
Counting is the family where the generative baseline is better, and the reason is specific to the
Score design. A count is a bounded measurement whose annotation is a single integer; \medjev{}
represents it as a distribution over an ordered grid and folds it to the expectation, while the baseline emits the integer directly, and its point estimates are closer on average, the only task-level difference distinguishable from zero at $95\%$ (Table~\ref{tab:pred}). The likely mechanism is the folding: an expectation over a wide grid shrinks extreme values towards its middle, so a model uncertain across the upper range reports a value closer to the centre. We state this as a hypothesis about the Score representation as implemented here; it is the only family in the suite where the probability object costs accuracy, and the mechanism is specific enough to be fixed rather than fundamental.

\section{Discussion}
\label{sec:disc}
\textbf{What the decision interface contributes.} The three decision types let heterogeneous medical outputs be expressed as explicit decisions: a class, a finding and a grid value are all candidate sets with a distribution over them, so the diagnosis, the findings and the counts in one report are probabilities of the same kind. The experiments show that these probabilities correspond far more closely to observed correctness, by an order of magnitude or more in three of the four families, and that they support selective behaviour the reconstruction cannot; no advantage in prediction is claimed, and none is needed. \textbf{\medjev{} does not primarily make the model more accurate; it makes the model faithful about how likely its decisions are to be correct.} The interface adds no information; it changes the object the model is trained to report, and that object is the one a decision rule consumes.

\textbf{What medicine adds to the idea.} Applying a decision interface in medicine is not a matter of porting one: a generalist model has to accept different modalities, cover different tasks, and express them in one comparable probability object, which is what we mean by omni. Two design commitments follow: several decisions can share one state, and candidate semantics belong in the input rather than in a fixed output layer. The contribution is therefore medical rather than architectural: the interface had to be extended to cover modalities and tasks, and medicine is where that extension is load-bearing.

\textbf{The scope of the claim.} The comparison cannot separate the interface from the denser supervision and the additional decision head that come with it, so what we measure is decision modeling as implemented rather than the interface alone. The probabilities are evaluated retrospectively on held-out data from the same sources as training, and per-class, per-source and out-of-distribution behaviour is not measured. Post-hoc calibration remains complementary: it repairs a number the model already emits, whereas the interface makes the decision an input.

\section{Conclusion}
\label{sec:conclusion}
We asked whether a generalist medical multimodal model can report confidence that corresponds to how often it is right, and whether making the decision its output object is the way to achieve it. Under an interface-controlled comparison the answer is yes on both counts: the reported probabilities track observed correctness far more closely, by up to two orders of magnitude in the reliability term, while point-prediction performance stays comparable and counting remains the boundary case. Generalist medical models should therefore be omni not only in what they perceive and solve, but also in how faithfully they quantify their own decisions.

\bibliographystyle{plainnat} \bibliography{references} \section*{Ethics Statement} This work uses only de-identified, publicly available medical imaging datasets and collects no new patient data. The models studied are retrospective predictors, not clinical tools: no component of this work is intended or validated for diagnostic use, and the decision probabilities reported here have not been assessed prospectively in a clinical workflow. The one place where a deployment claim could be read into the results is selective prediction, where a deferral policy consumes the model's confidence; we note explicitly that a model that ranks decisions well is not thereby safe to deploy, and that the human factors of deferral remain an open question (Appendix~\ref{app:related}). \section*{Reproducibility Statement} Both models train on public datasets listed in Appendix~\ref{app:suite}, with splits and candidate grids described there. Metric definitions and the estimator used to recover probabilities from the generative baseline are given in Sections~\ref{sec:exp-metrics} and~\ref{sec:exp-recover}, with the full definitions, edge cases and checks in Appendix~\ref{app:metrics} and~\ref{app:recover}. Training hyperparameters, hardware and checkpoint selection are in Appendix~\ref{app:train}; the per-checkpoint numbers underlying every figure are in Appendix~\ref{app:full}. Prompt templates for both models are reproduced in Appendix~\ref{app:prompts}.
\appendix
\section{Evaluation Suite Details}
\label{app:suite}

\begin{table}[t]
  \centering
  \caption{The evaluation suite. Both models train on the same 23{,}558 states and are
  evaluated on the same 726 held-out states. ``Decisions'' counts decision-level targets
  in training (Section~\ref{sec:method-many}); ``test'' counts held-out states. Candidate
  grid sizes are given for Score decisions and per-state candidate counts for Noul.}
  \label{tab:suite}
  \footnotesize
  \setlength{\tabcolsep}{3pt}
  \resizebox{\textwidth}{!}{%
  \begin{tabular}{@{}>{\raggedright\arraybackslash}p{2.0cm}>{\raggedright\arraybackslash}p{3.85cm}>{\raggedright\arraybackslash}p{2.0cm}>{\raggedright\arraybackslash}p{1.05cm}rrr@{}}
    \toprule
    Task family & Source datasets & Modalities & Type & States & Dec. & Test \\
    \midrule
    Classification & TissueMNIST, PathMNIST, OCTMNIST, Organ\{A,S,C\}MNIST, BloodMNIST, DermaMNIST, PneumoniaMNIST, BreastMNIST & Histology, Pathology, CT, Hematology, Dermatology, Ophthalmology & Choice & 10{,}000 & 10{,}000 & 200 \\
    \midrule
    Multi-label & ChestMNIST, ChestX-Det & Chest radiography & Noul $\times$14 / $\times$13 & 10{,}000 & 139{,}537 & 200 \\
    \midrule
    Counting & TXL-PBC & Haematology microscopy & Score $K{=}33$ & 878 & 3{,}512 & 126 \\
    \midrule
    Regression & CAMUS, RetinaMNIST & Ultrasound, Fundus & Score $K{=}101$ / $K{=}5$ & 2{,}680 & 2{,}680 & 200 \\
    \midrule
    \textbf{Total} & 15 datasets & 8 modality labels & 3 types & \textbf{23{,}558} & \textbf{155{,}729} & \textbf{726} \\
    \bottomrule
  \end{tabular}%
  }
\end{table}

\subsection{Composition and supervision}
Both models train on the same 23{,}558 states and are scored on the same 726 held-out states,
so the comparison holds at the level of states rather than of images. The suite spans fifteen
source datasets and eight modality labels, and the models differ in how those states become
targets. A generative baseline emits
one answer per state, so a chest radiograph with fourteen annotated findings contributes one
string; the decision interface factorizes the same annotation into one target per decision, so
the same radiograph contributes fourteen presence decisions, and the suite as a whole yields
155{,}729 decision-level targets rather than 23{,}558 answer-level targets. No additional
annotation is collected: the counts differ because the annotation is read at the level of the
decisions it describes. Table~\ref{tab:suite} gives the family-level totals, and the
per-source-dataset composition and splits are recorded in the evaluation artifacts.

\section{Typed Decision Spaces and Candidate Grids}
\label{app:decisions}

\subsection{Decision types and their signatures}
A decision is a tuple $D=(Q,T,C)$: a question $Q$, a type $T$, and a finite candidate
space $C=\{c_1,\dots,c_K\}$. The candidate space is part of the model input, not a fixed
output head, and the model returns a normalised vector over it. Four types are defined in the interface contract; only three of them are evaluated here.

\begin{itemize}
  \item \textbf{Choice.} $C$ is an unordered finite set of alternatives;
  $p_\theta(c_k \mid S,D) \ge 0$ and $\sum_{k=1}^{K} p_\theta(c_k \mid S,D) = 1$. The folded
  prediction is $\arg\max_k p_\theta(c_k \mid S,D)$. Used for classification and for the
  counting grid.
  \item \textbf{Noul.} $C = \{\text{No}, \text{Yes}\}$ and
  $p(\text{No}) + p(\text{Yes}) = 1$. One Noul decision is emitted per finding, so a single
  chest radiograph state produces 14 (chestmnist) or 13 (ChestX-Det) binary decisions in one
  forward pass. The \texttt{No} candidate is the null answer for that finding: it asserts
  absence rather than declining to answer. Folding keeps a label iff $p(\text{Yes}) > 0.5$
  and joins the surviving labels with ``;''.
  \item \textbf{Score.} $C=\{v_1,\dots,v_K\}$ is an ordered grid; the model returns
  $p(v_k)$ and the point estimate is the expectation $\hat{y} = \sum_k p(v_k)\, v_k$. Used
  for regression, and, with rounding, for the count decision. Candidates are not numbered:
  the grid position defines the index and the candidate text is self-contained
  (\texttt{- 45 >>} already says 45), so moving a candidate changes only the position it
  occupies.
  \item \textbf{Span.} The detection type, in which a box is emitted as a span decision.
  Span decisions were \emph{removed on both models}; no result reported here scores detection. The reason is recorded in the project log: the coordinate head collapsed
  to a constant box rather than localizing findings, with near-zero overlap against the
  annotations.
\end{itemize}

\subsection{Candidate grids}
Grid sizes follow the annotation space of each source dataset rather than the model.
Classification candidates are the pooled organ and tissue classes of the ten MedMNIST
collections. Noul decisions use two candidates per finding, with thirteen or fourteen decisions
per chest radiograph. Score decisions use an ordered grid: the platelet and white-cell counts are
represented on a 33-point grid, the ejection fraction on a 101-point grid, and the retinal score
on a 5-point grid. Candidates carry no injected index, so a Score grid is identified by the
values it contains and can be replaced at inference time without changing the model.

\subsection{Open-set behaviour of the null answer}
The \texttt{add\_none} probe inserts an explicit \texttt{none} candidate into every decision
space and measures whether its probability carries error information. At step 2424 the mean
$p(\texttt{none})$ is $0.3680$, and it is higher on correct decisions than on wrong ones
($0.4017$ versus $0.1697$); at step 1616 the same pattern holds ($0.3726$ overall, $0.4046$ on
correct versus $0.1858$ on wrong). The resulting AUC of $p(\texttt{none})$ as a predictor of
error is $0.1335$ at step 2424 and $0.1319$ at step 1616, i.e.\ reversed relative to the
intended direction. The project log records this failure as unexplained; we report it rather
than omit it.

\section{Metrics: Definitions, Motivation and Scope}
\label{app:metrics}

Every quantity used in this report is produced by one pure-function metric library, so a metric
means the same thing on both sides of the comparison. This appendix states the definitions, the
reason each family is reported, and what each family can and cannot establish.

\subsection{What each family can and cannot establish}
ECE is an average of absolute bin gaps, so it is comparable across models but insensitive to the
direction of the error, and it can hide a model that is overconfident in one region of
confidence space and underconfident in another; MCE reports the worst bin and is the reason we
do not read ECE alone. Reliability and resolution are reported together because either one in
isolation is uninformative: a model that always emits the class prior has reliability error
near zero and resolution exactly zero, and a model that ranks well but reports unusable numbers
has high resolution and high reliability error. The pair is what separates the two failure
modes, and it is the reason our central claim is about reliability rather than about accuracy.
AURC and $\mathrm{risk}@k$ are the quantities a deferral policy actually consumes, but they
measure the quality of an ordering, not its safety: a model can have a favourable AURC and
still be miscalibrated for a subpopulation, which is the failure mode the clinical
learning-to-defer literature reports. $\mathrm{DDx}@0.9$ and top-$k$ coverage are task-level
summaries of how the probability mass is distributed over the candidate set; they are
interpretable but threshold-dependent, and they inherit whatever error the candidate grid
introduces. Normalized entropy describes the shape of a decision distribution and is reported
for descriptive purposes only. Wilson intervals are used for proportions because several of
these quantities sit near $0$ or $1$ where the normal approximation fails, and paired bootstrap
intervals are used for every model-to-model difference because the two models score the same rows.

\subsection{The counting comparison on matched decisions}

Counting is the one family whose decision-level count differs between models: the decision
's counting metrics aggregate 201 decisions (the total-count decision of each of 126
states plus 75 cell-type counts), while the counting MAE and every generative-baseline number are
computed over the 126 total-count rows. Table~\ref{tab:appmatched} repeats the \medjev{}
numbers on the matched 126 decisions.

\begin{table}[t]
  \centering
  \caption{The counting family on the 126 total-count decisions both models scored. The upper
  block is the \medjev's own distribution; the lower block is the baseline's
  recorded values for the same rows. Values in parentheses are the recorded 201-decision
  figures that the main text reports for \medjev.}
  \label{tab:appmatched}
  \footnotesize
  \setlength{\tabcolsep}{5pt}
  \begin{tabular}{@{}llrr@{}}
    \toprule
    Arm & Quantity & 126 matched & (201 recorded) \\
    \midrule
    \medjev & top-1 (argmax) accuracy $\up$ & 0.1508 & (0.3234) \\
    \medjev & MAE of folded prediction $\dv$ & 2.5159 & (2.5159) \\
    \medjev & MAE of the expectation $\dv$ & 2.5179 & --- \\
    \medjev & RMSE of the expectation $\dv$ & 3.3250 & (3.3747) \\
    \medjev & mean maximum probability & 0.1811 & (0.3309) \\
    \medjev & median maximum probability & 0.1697 & (0.1889) \\
    \medjev & top-3 coverage of the gold count $\up$ & 0.4048 & (0.5473) \\
    \medjev & mean expectation minus mean gold & $-1.1252$ & --- \\
    \midrule
    \gen & exact integer match of the generated string $\up$ & 0.1349 & --- \\
    \gen & MAE of the generated integer $\dv$ & 2.1429 & --- \\
    \gen & top-1 of the reconstructed distribution $\up$ & 0.0000 & --- \\
    \gen & top-3 coverage of the reconstruction $\up$ & 0.3254 & --- \\
    \bottomrule
  \end{tabular}
\end{table}

Two things follow. First, the recorded counting calibration figures are optimistic for the
total-count task, because the extra 75 decisions are cell-type counts that the model finds
nearly deterministic: the white-cell count has a mean maximum probability of $0.9746$, and
including it lifts the family's mean maximum probability from $0.1811$ to $0.3309$ and top-3
coverage from $0.4048$ to $0.5473$. Second, the \medjev's top-1 accuracy on the
total-count task is $0.1508$ rather than the $0.3234$ the family aggregate reports, which is
the same effect seen from the accuracy side. Neither correction changes the direction of the
counting comparison: the generative baseline is ahead on the task metric, and \medjev is
far better calibrated on the same rows.

\section{Full Results Tables}
\label{app:full}

This appendix reports the per-checkpoint numbers behind Figures~\ref{fig:dynamics} and~\ref{fig:risk} and Tables~\ref{tab:pred} and~\ref{tab:quality}. Both models are evaluated at the same step, and the \medjev's values are recorded at every checkpoint we probed, while the baseline's decision-quality numbers exist only at the matched step.

\subsection{Task metrics at all ten checkpoints}
Table~\ref{tab:appdyn} lists the four task metrics for both models at all ten saved
checkpoints, plus the untuned backbone. The backbone row is a single evaluation of
MedGemma-1.5-4B before training, so it is shared by both models; it was run on the 1{,}126-row
file whose four-family subset is the 726-row test set, which makes its four-family numbers
directly comparable with the rows below.

\begin{table}[t]
  \centering
  \caption{Task metrics at all ten matched checkpoints. Classification is balanced accuracy
  ($\uparrow$), multi-label is F1 ($\uparrow$), counting and regression are MAE ($\downarrow$).
  Step 0 is the untuned backbone, evaluated once and therefore identical for both models; its
  counting and regression MAE count unparseable generated answers as errors ($84.5\%$ of its
  regression answers parsed), so those two cells are pessimistic. The
  selected comparison checkpoint is 2424 (in bold). These are the numbers plotted in
  Figure~\ref{fig:dynamics}.}
  \label{tab:appdyn}
  \footnotesize
  \begin{tabular}{@{}r rr rr rr rr@{}}
    \toprule
     & \multicolumn{2}{c}{Classification} & \multicolumn{2}{c}{Multi-label}
     & \multicolumn{2}{c}{Counting} & \multicolumn{2}{c}{Regression} \\
    \cmidrule(lr){2-3}\cmidrule(lr){4-5}\cmidrule(lr){6-7}\cmidrule(lr){8-9}
    Step & J & S & J & S & J & S & J & S \\
    \midrule
    0 (base) & 0.0000 & 0.0000 & 0.1813 & 0.1813 & 13.9127 & 13.9127 & 17.7775 & 17.7775 \\
    404  & 0.3709 & 0.4171 & 0.5283 & 0.4483 & 4.1905 & 3.9444 & 3.4214 & 3.7350 \\
    808  & 0.4021 & 0.4416 & 0.5553 & 0.5650 & 10.7698 & 3.9603 & 3.9326 & 11.5100 \\
    1212 & 0.5114 & 0.5345 & 0.5672 & 0.5100 & 3.5635 & 2.9048 & 3.8455 & 3.2500 \\
    1616 & 0.6275 & 0.6499 & 0.5902 & 0.5400 & 3.2222 & 2.7937 & 3.3723 & 3.3800 \\
    2020 & 0.6855 & 0.6379 & 0.5590 & 0.5683 & 2.8651 & 2.7222 & 3.5356 & 3.1150 \\
    \textbf{2424} & \textbf{0.6892} & 0.6469 & \textbf{0.5618} & 0.5449 & 2.5159 & \textbf{2.1429} & \textbf{2.9488} & 3.3400 \\
    2828 & 0.7049 & 0.6680 & 0.5778 & 0.5661 & 2.4683 & 2.0079 & 3.1090 & 2.9750 \\
    3232 & 0.7105 & 0.6940 & 0.5587 & 0.5549 & 2.3968 & 1.9286 & 3.0091 & 3.6000 \\
    3636 & 0.7141 & 0.7053 & 0.5412 & 0.5479 & 2.4365 & 1.7778 & 3.0175 & 3.0850 \\
    3685 & 0.7141 & 0.7022 & 0.5345 & 0.5529 & 2.4444 & 1.7857 & 3.0159 & 3.2800 \\
    \bottomrule
  \end{tabular}
\end{table}

Four features of Table~\ref{tab:appdyn} are load-bearing for the main text. (i) In
classification the two curves cross between steps 1616 and 2020: the generative baseline leads by
$+0.046$ at 404 and $+0.022$ at 1616, \medjev leads by $+0.048$ at 2020 and $+0.042$
at 2424. (ii) In counting the generative baseline wins every one of the ten steps and the gap
widens from $0.25$ at 404 to $0.37$ at 2424 and $0.66$ at the final step; the \medjev's
grid distribution is comparatively flat, and its folding rule takes the expectation
$\sum_k k\,p_k$, which regresses extreme counts toward the centre. (iii) In regression the
generative baseline has one severe mid-training spike (MAE $11.5100$ at 808) and a smaller rebound
at 3232 ($3.6000$, against $2.9750$ at 2828) while the
\medjev{} stays within $2.95$--$3.93$ throughout. (iv) In multi-label \medjev peaks
at $0.5902$ at step 1616, above the baseline's best value anywhere in training
($0.5683$ at 2020), and then declines to $0.5618$ at 2424. The trivial empty-prediction
baseline on this family is $0.5200$, and the \medjev's empty-prediction count at 2424 is
105 of 200 rows, against 190 of 200 at step 404, so the $0.5618$ is not produced by retreating
into silence.

\sloppypar
The supervision-density asymmetry of Appendix~\ref{app:suite} is the most plausible
explanation for (i): \medjev receives 155{,}729 decision-level signals from the same
23{,}558 images, of which 139{,}537 are multi-label Noul decisions, i.e.\ a large amount of
``read two candidate strings and choose one'' training that transfers to classification
Choice decisions, whereas the generative baseline receives one answer-level signal per row. We
state this as a hypothesis, not a finding; the ablation that would test it (down-sampling the
multi-label decisions to the baseline's count and re-measuring the crossing point) has
not been run.

\begin{table}[t]
  \centering
  \caption{Dynamic decision probes on the held-out states. Permutation is evaluated on the
  subset with more than one candidate; restriction and gold removal are evaluated on
  decisions whose annotated candidate is present. Chance levels are $0.50$ for a two-option
  shortlist and $0.25$ for a four-option shortlist; $0.5$ is the order-independence value
  for mean selected rank. The last three rows give total variation / top-1 agreement per
  family. The two columns are two checkpoints of the same model. These operations are
  not defined under the compared generative interface without
  an additional candidate-scoring mechanism (Section~\ref{sec:res-probes}).}
  \label{tab:probes}
  \footnotesize
  \setlength{\tabcolsep}{6pt}
  \begin{tabular}{@{}lrr@{}}
    \toprule
    Probe & @ 1616 & @ 2424 \\
    \midrule
    Permutation, total variation $\dv$ & 0.1284 & 0.1413 \\
    Permutation, top-1 agreement $\up$ & 0.6589 & 0.6622 \\
    Permutation, mean selected rank & 0.5486 & 0.5346 \\
    \quad diagnosis & 0.049 / 0.910 & 0.049 / 0.900 \\
    \quad counting & 0.155 / 0.373 & 0.198 / 0.418 \\
    \quad regression & 0.181 / 0.695 & 0.176 / 0.670 \\
    Restriction to 2 candidates $\up$ (chance 0.50) & 0.8735 & 0.8719 \\
    Restriction to 4 candidates $\up$ (chance 0.25) & 0.6717 & 0.7322 \\
    Gold removal, runner-up selected $\up$ & 0.8885 & 0.8752 \\
    \bottomrule
  \end{tabular}
\end{table}

\subsection{Decision-quality metrics at the probed checkpoints}
Decision-quality extraction was run at four checkpoints. On \medjev all four are
available; on the generative baseline the logit extraction was run only at step 2424, so the
generative column in Table~\ref{tab:appdyneq} has a single entry. Comparing the four
\medjev{} columns is therefore informative, while the generative column is a single
reference point rather than a trajectory.

\begin{table}[t]
  \centering
  \caption{Decision-quality metrics at the four probed checkpoints. Columns 1--4 are
  OmniMed-Jev; the last column is the baseline's reconstructed distribution, which was
  extracted only at step 2424. ECE, reliability and AURC are $\downarrow$; resolution is
  $\uparrow$; accuracy is the top-1 accuracy of the scored distribution.}
  \label{tab:appdyneq}
  \footnotesize
  \begin{tabular}{@{}l l r r r r r@{}}
    \toprule
    Family & Metric & 404 & 1616 & 2424 & 3685 & SFT @2424 \\
    \midrule
    \multirow{5}{*}{Class.}
      & Accuracy    & 0.3500 & 0.5700 & 0.6400 & 0.6750 & 0.5500 \\
      & ECE         & 0.1538 & 0.0863 & 0.0806 & 0.1310 & 0.2589 \\
      & Reliability & 0.0336 & 0.0123 & 0.0110 & 0.0310 & 0.1002 \\
      & Resolution  & 0.0602 & 0.1026 & 0.0860 & 0.0827 & 0.0938 \\
      & AURC        & 0.4354 & 0.1714 & 0.1351 & 0.1181 & 0.2020 \\
    \midrule
    \multirow{5}{*}{Multi-lab.}
      & Accuracy    & 0.9417 & 0.9420 & 0.9352 & 0.9338 & 0.5166 \\
      & ECE         & 0.0204 & 0.0204 & 0.0114 & 0.0216 & 0.2981 \\
      & Reliability & 0.0014 & 0.0010 & 0.0010 & 0.0006 & 0.1294 \\
      & Resolution  & 0.0087 & 0.0069 & 0.0102 & 0.0090 & 0.0776 \\
      & AURC        & 0.0167 & 0.0121 & 0.0126 & 0.0160 & 0.2474 \\
    \midrule
    \multirow{5}{*}{Count.}
      & Accuracy    & 0.1443 & 0.3085 & 0.3234 & 0.2886 & 0.0000 \\
      & ECE         & 0.0536 & 0.0538 & 0.0408 & 0.1792 & 0.4204 \\
      & Reliability & 0.0098 & 0.0099 & 0.0041 & 0.0423 & 0.1824 \\
      & Resolution  & 0.0127 & 0.1168 & 0.1087 & 0.1258 & 0.0000 \\
      & AURC        & 0.8213 & 0.3995 & 0.3890 & 0.4151 & 1.0000 \\
    \midrule
    \multirow{5}{*}{Regr.}
      & Accuracy    & 0.4700 & 0.4500 & 0.4800 & 0.4750 & 0.4650 \\
      & ECE         & 0.1121 & 0.0830 & 0.0913 & 0.1899 & 0.1267 \\
      & Reliability & 0.0198 & 0.0097 & 0.0205 & 0.0614 & 0.0258 \\
      & Resolution  & 0.0888 & 0.1299 & 0.1268 & 0.1233 & 0.1426 \\
      & AURC        & 0.3521 & 0.2504 & 0.2280 & 0.2393 & 0.2272 \\
    \bottomrule
  \end{tabular}
\end{table}

Three patterns are worth recording. The \medjev's counting resolution jumps by an order
of magnitude between 404 and 1616 ($0.0127 \to 0.1168$) and then stays near $0.11$--$0.13$,
while its counting ECE is worst at 3685 ($0.1792$). Its multi-label reliability falls
to $0.0006$ at 3685, the best value anywhere in the run, even though the
multi-label F1 at that step is its worst ($0.5345$). Regression calibration degrades after
2424 while its accuracy is flat, which is the reason 2424 and not 3685 is the comparison
checkpoint.

\section{Reproducibility, Compute and Statistical Protocol}
\label{app:train}

\subsection{Two-model design}
The comparison holds the backbone, the training images, the optimisation budget and the
option-order augmentation fixed, and changes only the output interface. Both models start from
\texttt{models/\allowbreak medgemma-1.5-4b-it}, see the same 23{,}558 images, take the same 3{,}685
optimisation steps, use the same seed, and are scored on the same 726-row test file by the
same evaluator. Table~\ref{tab:apphyper} lists the configuration item by item; exactly one
item differs.

\begin{table}[t]
  \centering
  \caption{Training configuration for the two models. Every row is identical on both sides
  except \texttt{max\_length}, which is a consequence of the interface: \medjev packs
  all of a state's decisions into one sequence, and the longest observed decision block is
  2{,}707 tokens, which does not fit in 1{,}024. This is not a tuning choice and is the only
  configuration difference between the models.}
  \label{tab:apphyper}
  \footnotesize
  \begin{tabularx}{\linewidth}{@{}>{\raggedright\arraybackslash}X l l@{}}
    \toprule
    Item & Generative SFT & OmniMed-Jev \\
    \midrule
    Backbone & \texttt{models/\allowbreak medgemma-1.5-4b-it} & same \\
    Quantisation & 4-bit NF4, double quantisation, bf16 compute & same \\
    LoRA rank / alpha / dropout & 16 / 16 / 0.05 & same \\
    LoRA target modules & \texttt{all-linear} (including the vision tower) & same \\
    Learning rate & $2\times10^{-4}$ & same \\
    Schedule & linear & same \\
    Warmup & 111 steps (ratio 0.03) & same \\
    Optimiser & \texttt{paged\_adamw\_8bit} & same \\
    Weight decay / gradient clipping & 0.0 / 0.3 & same \\
    Precision & bf16 & same \\
    Batch & $8 \times 4 = 32$ effective & same \\
    Epochs & 5 & 5 \\
    Total steps & 3{,}685 & 3{,}685 \\
    Sampling & \texttt{group\_by\_length} & same \\
    Seed & 42 & same \\
    Gradient checkpointing & on & on (Stage B) \\
    Option-order augmentation & \texttt{shuffle\_options: true} & \texttt{shuffle\_candidates: true} \\
    Checkpoint saving & \texttt{save\_steps = 404} (10 checkpoints) & same \\
    \midrule
    \textbf{\texttt{max\_length}} & \textbf{1{,}024} & \textbf{3{,}072} \\
    \bottomrule
  \end{tabularx}
\end{table}

Augmentation is applied during training and switched off at evaluation time, because the
development loss has to be comparable across steps. On \medjev only Choice and Noul
candidates are permuted; Score grids are left in place because the position \emph{is} the
ordinal index, and permuting them would permute the measurement scale. On the generative baseline
only tasks carrying \texttt{choices} (classification and multi-label) are permuted, which is
the exact counterpart. The augmentation gives each state on average five distinct orderings
over training, out of $11!$ for the largest grid.

\subsection{Two-stage training of \medjev}
Stage A freezes the backbone and trains the decision head only, for 200 steps, with
1{,}057{,}410 trainable parameters. Stage B enables LoRA together with the head, giving
39{,}555{,}202 trainable parameters out of 2{,}529{,}778{,}162, i.e.\ 1.564\% of the model.

\subsection{Wall clock, hardware and memory}
\begin{table}[t]
  \centering
  \caption{Compute for the two models. Each model ran on one A800 on a different device. The peak
  memory figure is for the \medjev's Stage B.}
  \label{tab:appcompute}
  \footnotesize
  \begin{tabularx}{\linewidth}{@{}>{\raggedright\arraybackslash}X r r@{}}
    \toprule
    & Generative SFT & OmniMed-Jev \\
    \midrule
    Wall clock & 46{,}280\,s = 12.86\,h & 50{,}204\,s = 13.95\,h \\
    Final training loss & 0.5805 & 0.7807 \\
    Peak memory & --- & 17.18\,GB (Stage B) \\
    Device & GPU 2 (single A800) & GPU 0 (single A800) \\
    \bottomrule
  \end{tabularx}
\end{table}

\subsection{Statistical protocol}
Every reported difference is computed on the two models' per-row predictions, because both models
score the same rows and the unpaired comparison would overstate the uncertainty. Proportions are
reported with Wilson intervals, which stay inside $[0,1]$ where several of these quantities sit,
and differences are reported with paired bootstrap intervals.

\paragraph{Checkpoint} The comparison is made at step 2424, the \medjev's best-validation
checkpoint, and both models are evaluated at that same step.

\paragraph{Paired tests} Because both models score the same rows, the uncertainty that matters
is the paired one, and we computed it on the two models' per-row predictions rather than
leaving the comparison at unpaired standard errors.

\begin{itemize}\itemsep2pt\sloppy
\item \emph{Classification balanced accuracy} differs by $+0.0422$ with a paired bootstrap
95\% interval of $[-0.0196, +0.1059]$. The $2\times2$ table over the 200 rows is: both
correct 109, \medjev{} only 19, generative baseline only 18, both wrong 54. The exact McNemar
test on the 37 discordant pairs gives $p = 1.00$. The unpaired standard-error statement
above ($1.2$ SE) therefore overstates how much the data support this difference: paired, the
evidence is essentially absent.
\item \emph{Multi-label macro F1} differs by $+0.0170$ with interval $[-0.0264, +0.0620]$,
and the per-row sign test gives 19 rows to \medjev, 23 to the generative baseline and
158 ties ($p = 0.64$).
\item \emph{Counting MAE} on the 126 matched rows differs by $+0.3730$ with interval
$[+0.0952, +0.6429]$. This is the only task-level difference that is distinguishable from
zero at 95\%, and it favours the generative baseline.
\item \emph{Regression MAE} differs by $-0.3912$ with interval $[-0.8756, +0.0173]$.
\end{itemize}

Bootstrap intervals use 20{,}000 paired resamples of rows with a freshly seeded generator
(seed 20260925) and are quoted to two decimals, the precision the Monte Carlo noise supports
($\pm 0.005$ at that budget). Nothing else in this appendix is a significance claim.

\section{Recovering Probabilities from the Generative Arm}
\label{app:recover}

\subsection{Why the generative baseline has to be reconstructed}
A generative model fine-tuned to emit only an answer does not report its own confidence because a sentence is appended at inference time. Table~\ref{tab:appsftprob} records the elicitation attempts we ran. Prompting for a distribution did not work in either of the two direct forms, and an untuned checkpoint produced values that were all zero, which is why every generative probability in this report comes from reading the model's own next-token logits. The failure is itself part of the result: obtaining confidence from a generative interface requires training it into the model, which is already a change of interface.

\begin{table}[t]
  \centering
  \caption{Attempts to elicit a distribution from the generative baseline. Attempts 1 and 2 failed
  outright; attempt 3 produced values but all of them were zero; attempt 4 is the estimator used
  for every generative number in this report.}
  \label{tab:appsftprob}
  \footnotesize
  \begin{tabularx}{\linewidth}{@{}c >{\raggedright\arraybackslash}X l@{}}
    \toprule
    & Estimator & Outcome \\
    \midrule
    1 & Append a \texttt{PROBS: \{\dots\}} request to the end of the user turn. & 0 of 726 rows parsed a confidence. \\
    2 & Write a two-line requirement into the system prompt (answer on line 1, confidence on line 2). & 0 of 20 rows parsed a confidence. \\
    3 & Untuned base model under the same instruction. & 8 of 20 rows produced a confidence and 1 of 20 produced scores, but every value was $0.0$. \\
    4 & Read the option log-likelihoods from the model's own logits. & Succeeded; the full metric set is computable. \\
    \bottomrule
  \end{tabularx}
\end{table}

\subsection{The estimator actually used}
For a candidate $c$ with continuation tokens $c_1,\dots,c_T$, the estimator scores each
candidate continuation separately and then normalises across candidates:
\begin{align}
  \log P(c) &= \sum_{t=1}^{T} \log p\!\left(c_t \mid \text{prompt}, \text{image}, c_{<t}\right),
  \label{eq:applogp}\\
  \tilde{p}(c) &= \operatorname*{softmax}_{c \in C} \log P(c) .
  \label{eq:appsoftmax}
\end{align}
Each candidate's continuation is prefixed with a space, i.e.\ in word-initial form. All $K$
candidates are scored in one batched forward pass. The multi-label family has no per-finding
likelihood in the semicolon-joined answer, so it is rewritten as a per-finding binary
question, ``Is \textit{<finding>} present in this image? Answer Yes or No.'', asked once per finding (14 times), and the softmax in Equation~\eqref{eq:appsoftmax} is taken over the
\texttt{Yes} and \texttt{No} continuations to give $p(\text{finding present})$.
Classification, counting and regression use the same prompt as the generative inference path.

\subsection{Two caveats on these numbers}
\emph{(i) The counting cell of the generative baseline is diagnostic, not decisive.} On the counting
family the top-1 candidate of the reconstructed distribution never equals the gold count:
$0$ of $126$ rows. That single fact mechanically forces resolution to exactly $0$ and AURC to
exactly $1.0$ for that cell, because a distribution whose top-1 is never correct cannot separate
easy from hard rows under a confidence ordering, and answering the most confident fraction of
decisions is then no better than answering all of them. The counting row of the generative baseline
should therefore be read as a statement about the reconstruction on a 33-way integer grid, not
as an estimate of how much confidence information a generative model carries in general.

\emph{(ii) The baseline's post-training is a strict superset of the \medjev's.} The
generative baseline was instruction-tuned after supervised fine-tuning, which is a strictly larger
post-training budget than \medjev received. If this asymmetry matters at all, it
favours the generative baseline, so the decision-quality gaps reported in
Appendix~\ref{app:metrics} are conservative with respect to it. We report the asymmetry rather
than leave it implicit.

\begin{sloppypar}
One further distinction belongs here. The distribution in
Equations~\eqref{eq:applogp}--\eqref{eq:appsoftmax} is a pseudo-distribution: nothing in
training ever made these log-likelihoods sum to one over the candidate set, and the
normalisation is ours, not the model's. The \medjev's distribution is the training
objective itself. The comparison is therefore between a native decision distribution and an
inference-time reconstruction of one, and the reconstruction is the more favourable of the two
constructions we could have chosen for the baseline.
\end{sloppypar}

\section{Prompt Templates}
\label{app:prompts}

Everything in this appendix is a verbatim rendering of the strings the models receive, not a paraphrase. Two editorial conventions apply. First, section headings and
parenthetical annotations that appear \emph{inside} the listings in the project log are
rendered in English here; every model-facing string is reproduced character for character.
Second, long lines are wrapped at spaces so that they fit the text block; no characters were
added, removed or reordered. The two decision markers are real single characters, U+27E9 (\texttt{>>} below; token id 241287, terminating a candidate) and U+2726 (\texttt{*} below; token id 244193, terminating a question), written here as ASCII stand-ins because the pdfLaTeX text fonts cannot typeset those code points.

\subsection{Task instructions shared by both models}
The same four blocks are used for training, for inference, and for the logit probe.

{\footnotesize\begin{verbatim}

[disease_diagnosis_classification]
You are given a medical image. Answer the classification question using only the provided
image. If options are provided, return only the correct option letter or class label.

[multi_label_classification]
You are given a medical image. Identify all applicable findings or labels. Return labels
separated by semicolons. Return an empty string if none apply.

[cell_counting]
You are given a medical image. Return only the integer count requested.

[regression]
You are given a medical image. Return only the requested numeric measurement.
\end{verbatim}}

The system prompt is identical on both models and in all three settings:
\texttt{You are an expert medical imaging assistant.}

\subsection{\medjev: the decision block}
The markers are the real single characters named above.

{\footnotesize\begin{verbatim}

<bos><start_of_turn>user
<start_of_image>[256 image soft tokens]<end_of_image>
{source} | {modality}

Q1: <question text verbatim>
- <candidate text> >>
- <candidate text> >>
*
Q2: ...
<end_of_turn>
\end{verbatim}}

One structural difference from the generative baseline deserves emphasis: \medjev does
\emph{not} use a \texttt{<start\_of\_turn>model} turn. The entire decision block sits inside the
user turn and ends with \texttt{add\_generation\_prompt=False}, because the model does not generate anything: it reads the hidden states at the marker positions. Readers who assume the
two sequence structures are the same will misread the interface comparison.

The four families as actually encoded (\texttt{...} marks elision):

{\footnotesize\begin{verbatim}

----- classification (organamnist, 11 candidates, 1 decision) -----
state header: organamnist | CT

Q1: Which class does this image belong to?
- bladder >>
- femur-left >>
- femur-right >>
- heart >>
- kidney-left >>
- kidney-right >>
- liver >>
- lung-left >>
- lung-right >>
- pancreas >>
- spleen >>
*

----- multi-label (ChestX-Det, 13 Noul decisions) -----
state header: ChestX-Det | Radiology

Q1: Is Atelectasis present in this Radiology image?
- No >>
- Yes >>
*
Q2: Is Calcification present in this Radiology image?
- No >>
- Yes >>
*
... (13 decisions in total)

----- counting (TXL-PBC, 4 Choice decisions, 33 candidates each) -----
state header: TXL-PBC | Hematology

Q1: How many cells are present in this blood smear image?
- 0 >>
- 1 >>
... (0...32)
- 32 >>
*
Q2: How many RBCs are present in this blood smear image?
... (likewise 0...32)
*
... (4 decisions in total: total + RBC / WBC / Platelets)

----- regression (CAMUS, 1 Score decision, 101 candidates) -----
state header: CAMUS | Cardiology

Q1: What is the left ventricular ejection fraction, in percent?
- 0 >>
- 1 >>
... (0...100)
- 100 >>
*
\end{verbatim}}

The family and type names in the four headers are printed by the routine that renders these
prompts, and one of them disagrees with the assignment the experiments use: the counting block is
labelled \texttt{4 Choice decisions} in the header, while the suite assigns counting a Score
decision over an ordered grid and scores it that way (Table~\ref{tab:suite}). The headers are
reproduced verbatim rather than corrected.

The key convention is that candidates carry no injected index: a Score grid is indexed by
position and the candidate text is self-contained (\texttt{- 45 >>} already says 45), so moving
a candidate changes only the position it occupies and never how it reads.

\section{Extended Related Work}
\label{app:related}
This appendix carries the threads that the main-text related work compresses, in the same
order: how medical calibration is measured, why the deficit is structural, what the
calibration-aware objectives do, and what the deferral, human-factors and regulatory
literature assumes about the probability it consumes. One thread is left to this appendix
because it is outside medicine: the decision-native wave now spans legal documents, video
quality, scientific relation choice, game control, edge service orchestration and population
experiments \citep{jevlegal2026,jevqa2026,jevscientific2026,jevstar2026,jevedge2026,jev6g2026,fractalmap2026,kite2026}.

\subsection{How medical calibration is measured}
Medical confidence is usually measured with the tools of probabilistic forecasting. The
Brier score and its Murphy decomposition into reliability, resolution and uncertainty
\citep{brier1950,murphy1973} separate a model's calibration from its discrimination, and
clinical reporting guidance insists on the distinction: discrimination measures such as
AUROC are invariant to monotone rescaling and therefore cannot substitute for a calibration
statement \citep{vanCalster2025metrics,banerji2023uncertainty}. Applied to medical
multimodal models, the measurement consistently finds overconfidence: hallucination and
overconfidence correlate across medical VQA benchmarks \citep{medhalluc2025}, calibration
degrades further under distribution shift \citep{clear2026}, per-sample calibration error
exposes cases that aggregate ECE hides \citep{persample2026}, and self-reported confidence is
a poor proxy for correctness in radiology and other specialties
\citep{omar2025benchmarking,boie2026selfreported}. A parallel line in cognitive science asks
whether models have any metacognitive access to their own errors at all, with results that
are mixed at best \citep{steyvers2025know,griot2025metacognition}. Our measurement protocol
follows this literature (top-label Murphy decomposition, AURC and risk--coverage) and adds
one object it does not usually have: a native decision distribution to measure.

\subsection{Why the deficit is structural}
The degradation is introduced by the post-training stack rather than by pretraining.
Instruction tuning raises confidence faster than accuracy \citep{xiong2024can,huang2024relying};
alignment and preference optimization can further distort the probability \citep{hu2026alignment};
temperature scaling recovers some of it but does not restore the ordering the model needs
\citep{xiao2025restoring}; and models report their own confidence inconsistently across
prompting and elicitation formats \citep{sanzguerrero2026own,wang2025objective}. Work on
learning to defer has long argued that a model's own probability is a poor basis for
allocation decisions \citep{mozannar2020consistent}, which is the assumption our interface
attacks at the source rather than after the fact.

\subsection{Calibration-aware objectives and their failure modes}
A large literature trains the probability instead of repairing it: teaching models to express
calibrated confidence in words or numbers \citep{lin2022teaching,kadavath2022know,
tian2023just}, semantic-entropy and distributional uncertainty over meaning
\citep{kuhn2023semantic,malinin2021uncertainty}, tokenized proper scoring rules and their
extensions \citep{confTuner2025,leng2025taming,gao2026capo,ma2026dcpo}, and reward shaping
that pays for honest uncertainty \citep{baniharouni2026rewarding,verma2023learning,
strong2025guided}. These methods work on the reward or the readout while leaving the output
interface generative, so the calibrated quantity is still a string; badly shaped confidence
rewards are additionally hackable \citep{rewardhack2026}, and calibration gains can be
bought at the cost of task performance or verifiability
\citep{verifiable2026,balancing2026}. The acronym RLCD also denotes reinforcement learning
from contrastive distillation \citep{rlcd2024}, an unrelated objective, and the two are kept
apart here. Our comparison differs in that the probability is the
model's output object from the first training step.

\subsection{Deferral, human factors and regulation}
Selective prediction and learning-to-defer methods consume a probability and decide whether
to act on it, including for medical tasks \citep{geifman2017selective,strong2025guided,
sun2026l2dguidance,vazhentsev2026abstention}. Three empirical findings constrain how much
that machinery can recover. First, abstention is nearly orthogonal to accuracy: models
rarely abstain where it matters clinically \citep{medabstain2026,medqabstain2026}. Second,
class-dependent miscalibration makes abstention actively harmful for under-represented
conditions even when aggregate calibration error looks small \citep{lopez2026selective}, and
conformal guarantees are weaker than they appear under clinical distribution shift
\citep{mehrtens2025pitfalls}. Third, deferral does not remove error but relocates it:
clinicians told that a model abstained miss more diagnoses and treatments than with no model
at all \citep{jabbour2025selective}, and cascading on the same confidence signal buys no
cost-efficiency on medical questions \citep{fanconi2025cascade}. Human-factors work
quantifies the stakes: biased or overconfident output degrades clinician accuracy even after
AI-literacy training \citep{qazi2026automation}, over-reliance persists when the advice is
wrong \citep{shekar2025overtrust,dratsch2023automation,jabbour2024measuring}, and uncertainty
expressed in words does not reliably reduce it
\citep{kim2024notsure,nicolson2025human,xu2026divergent,sikora2026idk}.

Evaluation and governance guidance now asks for exactly the probability this literature
cannot supply. Systematic reviews of clinical LLM evaluations find uncertainty rarely
reported \citep{bedi2025testing}; reporting standards make calibration and uncertainty
reporting explicit \citep{tripodllm2025,tripodai2024}; and regulators and professional bodies
expect calibrated uncertainty and deferral behaviour to be described per case
\citep{who2024lmm,fda2026generative,nist2024genai,ecri2026hazards,euAIAct2024}. Recent work
on clinical agents treats operational safety as a property of the whole system rather than
of the model \citep{saroc2026,zhang2026onpremise}, and evaluation-methodology work warns that
apparent gains can be artefacts of the protocol \citep{kapoor2024must}. Our study is written
to that standard: the comparison is interface-controlled, the uncertainty is reported with
paired intervals, and the places where the decision interface does not help are stated in the
results rather than deferred to a limitations paragraph.
\section{Limitations}
\label{app:limits}
\paragraph{Bounded decisions only} The interface covers decisions with enumerable candidate
sets, and this paper evaluates that scope: free-form report generation and coordinate
regression are not scored here. For detection we also have to report that our implementation
did not produce usable localization output, because the coordinate head collapsed to a constant
box (Appendix~\ref{app:suite}), so no claim is made about that family in either direction. A
medical model that must narrate remains a generative model for that part of its output; the
claim of this paper is about the bounded decisions that dominate structured clinical
prediction.

\paragraph{Factorized supervision} The two models see the same 23{,}558 images but the
decision interface factorizes their annotations into 155{,}729 decision targets
(Section~\ref{sec:method-many}). No additional annotation is collected, but the
supervision is denser, and denser supervision plausibly explains part of the late
classification crossover (Section~\ref{sec:res-pred}). Separating the interface effect
from the density effect would require an ablation outside this protocol, so the
crossover is reported as a hypothesis. The predictive differences themselves are small
relative to the test set: the classification interval $[-0.0196, +0.1059]$ contains zero,
which is why we claim a consistent direction across families rather than
significance at a point.

\paragraph{Evaluation scope} Four families, 726 held-out states and three decision types
are enough to test whether the interface effect is general, and not enough to establish
clinical utility. The probabilities reported here are model outputs on retrospective data;
no claim is made about prospective performance, workflow integration or patient outcomes,
and the human-factors evidence on deferral and over-reliance
\citep{jabbour2024measuring,dratsch2023automation,qazi2026automation,shekar2025overtrust,
kim2024notsure,nicolson2025human,xu2026divergent,sikora2026idk} is a reason to test those
questions rather than to assume them.

\end{document}